\documentclass[wcp]{jmlr}

\usepackage{longtable}

\usepackage{booktabs}

\usepackage{algorithm}
\usepackage{algorithmic}

\makeatletter
\let\Ginclude@graphics\@org@Ginclude@graphics 
\makeatother
\usepackage[load-configurations=version-1]{siunitx} 

\title[RDLDA]{Regularized Deep Linear Discriminant Analysis}

  \author{
  \Name{Wen Lu} \Email{wenlu@hbut.edu.cn}\\
   \addr Hubei University of Technology, China}

\begin{document}

\maketitle

\begin{abstract}
As a non-linear extension of the classic Linear Discriminant Analysis(LDA), Deep Linear Discriminant Analysis(DLDA) replaces the original Categorical Cross Entropy(CCE) loss function with eigenvalue-based loss function to make a deep neural network(DNN) able to learn linearly separable hidden representations. In this paper, we first point out DLDA focuses on training the cooperative discriminative ability of all the dimensions in the latent subspace, while put less emphasis on training the separable capacity of single dimension. To improve DLDA, a regularization method on within-class scatter matrix is proposed to strengthen the discriminative ability of each dimension, and also keep them complement each other. Experiment results on STL-10, CIFAR-10 and Pediatric Pneumonic Chest X-ray Dataset showed that our proposed regularization method Regularized Deep Linear Discriminant Analysis(RDLDA) outperformed DLDA and conventional neural network with CCE as objective. To further improve the discriminative ability of RDLDA in the local space, an algorithm named Subclass RDLDA is also proposed. 
\end{abstract}

\section{Introduction}

As a supervised dimensionality reduction and classification method, Linear Discriminant Analysis(LDA) projects the input data into a low dimensional subspace and finds the optimal linearly separable boundaries between classes in that subspace by maximizing the trace of between-class scatter matrix and minimizing the trace of within-class scatter matrix. LDA requires the input data be vectors, to process 2D data as images, the input data need to be vectorized. However, the dimensions of vectorized images usually exceed the sample size, which makes the within-class scatter matrix not invertible. Vectorization also requires more computational resources and loses the spatial information in 2D data. To solve the problem,  \cite{ye2004two}, \cite{li20052d}, \cite{xiong2005two} proposed matrix-based linear discriminant analysis(2D-LDA).

Since LDA is a linear transformation, it was born not good at processing data with nonlinear distribution. Inspired by Kernel PCA, nonlinear kernel methods are also introduced to LDA (\cite{mika1999fisher}), but those approaches are either limited by some fixed nonlinear transformations or have very complicated computation (\cite{li2019discriminant}). DNN has been widely recognized as a successful nonlinear representation learner, to utilize such talent of DNN,  \cite{dorfer2015deep} put LDA on top of DNN to form Deep Linear Discriminant Analysis(DLDA). Instead of maximizing the likelihood of target labels as conventional CCE, the eigenvalues along the discriminant eigenvector directions are maximized. By focusing on directions in the latent space with smallest discriminative power, DLDA learns linearly separable hidden representations with similar discriminative power in all directions of the latent space. Although \cite{dorfer2015deep} showed DLDA outperforms DNN with CCE under the same network architecture, there is still some improvement potential in DLDA.

In this paper, we first revisit the basic ideas of LDA, DLDA, autoencoder and K-means in Section \ref{section:Background}; then we explore the distribution pattern of DLDA hidden representations to find the improvement orientation in Section \ref{section:Distribution Pattern}. To improve DLDA, in Section \ref{section:RDLDA}, a regularization method on within-class scatter matrix is proposed to strengthen the discriminative ability of each latent subspace dimension. In Section \ref{section:Subclass_RDLDA}, we further propose an algorithm based on autoencoder and K-means, which splits each class into serveral subclasses and then trains the model to seperate all the subclasses in order to strengthen the discriminative ability in all the local space. In Section \ref{section:Experiments}, experiment results on the STL-10, CIFAR-10 and Pediatric Pneumonic Chest X-ray Dataset showed our proposed regularization method RDLDA outperformed DLDA and conventional neural network with CCE as objective. The effectiveness of our proposed algorithm was also verified by the CIFAR-10 Dataset experiments.

\section{Background} \label{section:Background} 

\subsection{Linear Discriminant Analysis}
As a supervised dimensionality reduction and classification method, Linear Discriminant Analysis(LDA) projects the input data into a low dimensional subspace and finds the optimal linearly separable boundaries between classes in that subspace.
Denote $\mathbb{R}^d\ni\{\boldsymbol{x}_i^{(j)}\}_{i=1}^{n_j}$ are the instances of the $j$-th class of which there are multiple number of classes ( \cite{ghojogh2019fisher} ). Between-class scatter matrix $\boldsymbol{S_B}$ and within-class scatter matrix $\boldsymbol{S_W}$ are introduced to measure the effectiveness of separation in the lower-dimensional subspace. 
The between-class scatter matrix is defined as:
\begin{equation}
\mathbb{R}^{d\times d}\ni \boldsymbol{S_B}:=\sum_{j=1}^cn_j(\boldsymbol{\mu}_j-\boldsymbol{\mu})(\boldsymbol{\mu}_j-\boldsymbol{\mu})^\top,
\end{equation}
where $c$ is the number of classes and:
\begin{equation}
\mathbb{R}^{d}\ni \boldsymbol{\mu}:=\frac{1}{\sum_{k=1}^cn_k}\sum_{j=1}^cn_j\boldsymbol{\mu}_j=\frac{1}{n}\sum_{i=1}^n\boldsymbol{x}_i
\end{equation}
is the weighted mean of means of classes or the total mean of data.
The within-class scatter matrix is defined as:
\begin{equation}
\mathbb{R}^{d\times d}\ni \boldsymbol{S_W}:=\sum_{j=1}^c\sum_{i=1}^{n_j}(\boldsymbol{x}_i^{(j)}-\boldsymbol{\mu}_j)(\boldsymbol{x}_i^{(j)}-\boldsymbol{\mu}_j)^\top,
\end{equation}
where $n_j$ is the sample size of the $j$-th class.

The target of LDA is to find the optimal linear projection directions, $\{\boldsymbol{u}_j\}_{j=1}^p$ where $\boldsymbol{u}_j\in \mathbb{R}^{d}$, that maximize the between-class scatter matrix $\boldsymbol{S_B}$ and minimize the within-class matrix $\boldsymbol{S_W}$, so the instances belong to different classes get far from each other while the instances of the same class get close to one another. The optimization problem is:
\begin{equation}
\max_{\boldsymbol{U}} f(\boldsymbol{U}):=\frac{\mathbf{tr}(\boldsymbol{U}^\top \boldsymbol{S_B}\boldsymbol{U})}{\mathbf{tr}(\boldsymbol{U}^\top \boldsymbol{S_W}\boldsymbol{U})},
\end{equation}
where $\mathbb{R}^{d\times p}\ni\ \boldsymbol{U}=[\boldsymbol{u}_1, \dots, \boldsymbol{u}_p]$.

The solution to the generalized Rayleigh-Ritz Quotient problem is:
\begin{equation}
\boldsymbol{U} = \mathbf{eig}(\boldsymbol{S_W}^{-1}\boldsymbol{S_B}),
\end{equation}
where $\mathbf{eig}(.)$ denotes the eigenvectors of the matrix stacked column-wise.

$\boldsymbol{S_W}$ might be singular and not invertible, especially in small sample size problems, where the number of available training samples is smaller than the dimensionality of the sample space. Meanwhile, the estimation of $\boldsymbol{S_W}$ overemphasizes high eigenvalues whereas small eigenvalues are underestimated. To make $\boldsymbol{S_W}$ invertible and correct the bias, a very small positive number $\lambda$ is added to the diagonal of $\boldsymbol{S_W}$  (\cite{friedman1989regularized}, \cite{lu2005regularization}, \cite{stuhlsatz2012feature}). In this case, the solution is:
\begin{equation}
\boldsymbol{U} = \mathbf{eig}((\boldsymbol{S_W}+\lambda \boldsymbol{I})^{-1}\boldsymbol{S_B}).
\end{equation}

\subsection{Deep Linear Discriminant Analysis}
Since each eigenvalue $v_i$ is the quantitative measurement of the discrimination ability in the direction of the corresponding eigenvector $\boldsymbol{u}_i$, as a non-linear extension of the classic LDA, Deep Linear Discriminant Analysis(DLDA) replaces the original Categorical Cross Entropy(CCE) loss function with eigenvalue-based loss function to encourage deep neural network(DNN) to learn feature representations with discriminative distribution parameters(\cite{dorfer2015deep}). To avoid DNN focusing on maximizing the eigenvalues of which the eigenvectors already separate the classes and ignoring the eigenvalues whose eigenvectors poorly discriminate the classes, the optimization object of DNN is limited to the smallest of all $c-1$ valid eigenvalues, so that all of the $c-1$ latent feature dimensions would be trained on balance. The target of DLDA is to find the optimal model parameters $\Theta$ of DNN such that the produced latent features have low intra-class variance and high inter-class variance:
\begin{equation}\label{eq:eigenvalue loss}
\mathop{\arg\max}_{\Theta}\frac{1}{k}\sum_{i=1}^kv_i \quad with \quad \{v_1, \dots, v_k\} = \{v_j|v_j<min\{v_1, \dots, v_{c-1}\}+\epsilon\}.
\end{equation}

\subsection{Autoencoder}
Autoencoder is a multi-layer neural network that attempts to reconstruct its input(\cite{ren2017autoencoder}). It consists of encoder layers, an embedding representation layer, and decoder layers. For reconstructing the input, the dimension of the output layer is the same as the input layer, and for extracting the discriminative hidden representation features from the input, the dimension of the embedding representation layer is smaller than the input layer. The simplest autoencoder network consists of a linear layer for encoder and a linear layer for decoder. The encoder layer maps the input sample $\boldsymbol{x}_i$ from the original data space to the hidden feature space $\boldsymbol{h}_i$ by:
\begin{equation}\label{eq:encoder}
\boldsymbol{h}_i = \sigma_1(\boldsymbol{W}\boldsymbol{x}_i + \boldsymbol{b}).
\end{equation}

The decoder maps the sample from the hidden feature space to the original data space by:
\begin{equation}\label{eq:decoder}
\boldsymbol{\hat{x}}_i = \sigma_2(\boldsymbol{W'}\boldsymbol{h}_i + \boldsymbol{b'}).
\end{equation}

In Eq.~\eqref{eq:encoder} and Eq.~\eqref{eq:decoder}, $\boldsymbol{W}$ and $\boldsymbol{W'}$ are the weight matrices , $\boldsymbol{b}$ and $\boldsymbol{b'}$ are the biases, and $\sigma_1$ and $\sigma_2$ are the activation functions, of the encoder layer and decoder layer respectively. The autoencoder network is trained to minimize the following reconstruction error as loss function:
\begin{equation}\label{eq:reconstruction_error}
Loss = \frac{1}{n}\sum_{i=1}^n \parallel \boldsymbol{x}_i - \boldsymbol{\hat{x}}_i \parallel ^2 = \frac{1}{n}\sum_{i=1}^n \parallel \boldsymbol{x}_i - \sigma_2(\boldsymbol{W'}\sigma_1(\boldsymbol{W}\boldsymbol{x}_i + \boldsymbol{b}) + \boldsymbol{b'}) \parallel ^2.
\end{equation}

\subsection{K-means clustering}
K-means is an unsupervised machine learning algorithm used to identify clusters of a dataset. The first step is to randomly generate K initial centroids, where K is the number of clusters desired. Each point is assigned to its closest centroid, and each collection of points assigned to the same centroid is a cluster. The centroid of each cluster is then updated based on the points assigned to the cluster. The iteration proceeds until the centroids remain the same.

\section{Distribution Pattern of DLDA Hidden Representations} \label{section:Distribution Pattern} 
\begin{figure}[htp]
\begin{center}
\includegraphics[width=\textwidth]{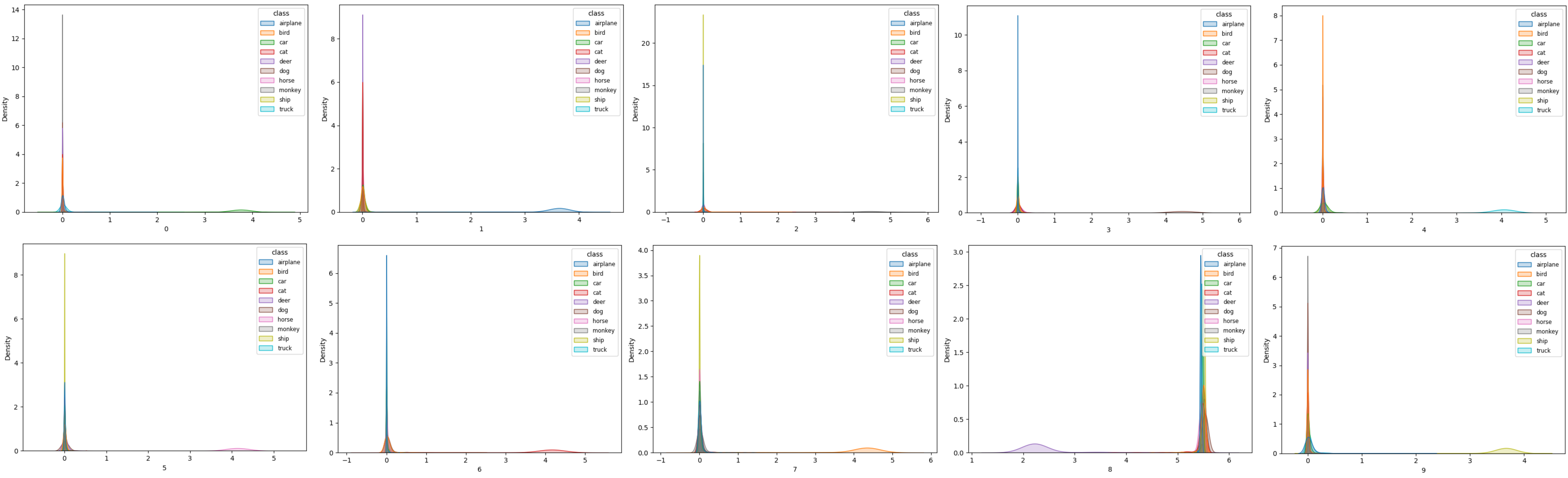}
\caption{Each subplot is the distribution of each output dimension of the DLDA DNN, which was trained 400 epochs by the STL-10 dataset.}\label{fig:STL10_train_output_1.0_kdeplot}
\end{center}
\end{figure}

To understand how a trained DLDA DNN nonlinearly `projects' the input data into a low dimensional subspace in which classes are linearly separable, the distribution of each output layer dimension of a DLDA DNN well-trained by the STL-10 dataset is displayed in each subplot of Fig.~\ref{fig:STL10_train_output_1.0_kdeplot} respectively(the network structure and training parameters setting are detailed in the experiments section of this paper). As seen from Fig.~\ref{fig:STL10_train_output_1.0_kdeplot}, the discriminative ability of each output dimension is so weak that it only separates a unique class, leaving the other nine classes clustered closely, but the ten dimensions cooperating all together are able to discriminate all the ten classes. This is a very interesting phenomenon, and another experiment on the CIFAR-10 dataset also shows analogous phenomenon.

\begin{figure}[htp]
\begin{center}
\includegraphics[width=\textwidth]{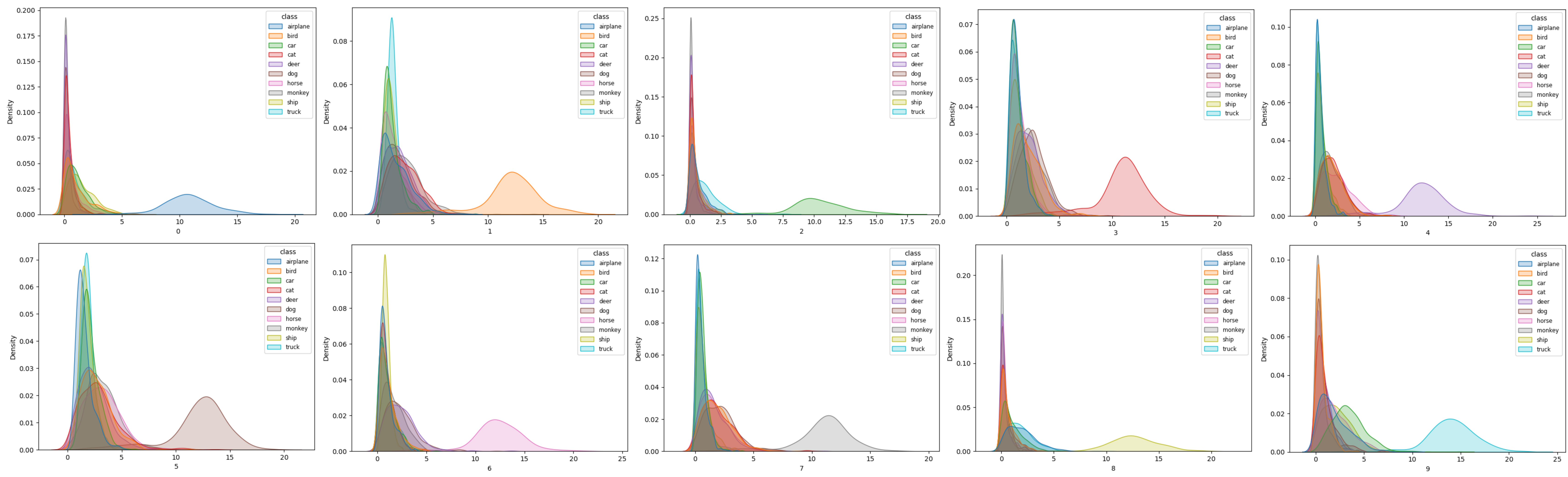}
\caption{Each subplot is the distribution of each output dimension of the CCE DNN, which was trained 400 epochs by the STL-10 dataset.}\label{fig:STL10_train_DNN_kdeplot}
\end{center}
\end{figure}

This distribution pattern coincides with that of the conventional neural network with CCE as objective, as shown in Fig.~\ref{fig:STL10_train_DNN_kdeplot}, which may not be a surprise. One of the two differences is that the discriminated class is fixed for each output dimension of CCE DNN, while it is not fixed for DLDA DNN. For a certain dimension, different runs might separate a different class. The other difference is that the distribution variances of DLDA DNN output dimensions are much smaller than those of CCE DNN output dimensions, which may explain why DLDA DNN performs better than CCE DNN in the experiments of Section \ref{section:Experiments}.

Although technically the output dimensions cannot be used directly for classification, the experiments results in Tab.~\ref{table:STL-10 experiment results} and Tab.~\ref{table:CIFAR-10 experiment results} of Section \ref{section:Experiments} showed the following three predictors built upon the output dimensions(hidden representations) of DLDA DNN have nearly identical prediction accuracy.

{\bfseries Hyperplanes:}
\cite{dorfer2015deep} proposed to rely on the distances to the linear decision hyperplanes (\cite{friedman2001elements}) for classification.  $\boldsymbol{e} = \{ \boldsymbol{e}_i\}_{i=1}^{c-1}$ are the corresponding eigenvectors, based on the LDA projection matrix $\boldsymbol{A}:=[\boldsymbol{e}_1, \dots, \boldsymbol{e}_{c-1}]$ and the per-class mean hidden representations $\bar{\boldsymbol{H}}_c:=[\boldsymbol{\bar{h}}_1^\top, \dots, \boldsymbol{\bar{h}}_c^\top]$, the distances of test sample hidden representation $\boldsymbol{h}_t$ to the linear decision hyperplanes are defined as:
\begin{equation}
\boldsymbol{d} = \boldsymbol{h}_t^\top \boldsymbol{T}^\top -\frac{1}{2}\mathbf{diag}(\bar{\boldsymbol{H}}_c\boldsymbol{T}^\top) \quad with \quad  \boldsymbol{T}=\bar{\boldsymbol{H}}_c\boldsymbol{A}\boldsymbol{A}^\top
\end{equation}
where $\boldsymbol{T}$ are the decision hyperplane normal vectors. The class probabilities vector is $\boldsymbol{p}'_c=1/(1+e^{-\boldsymbol{d}})$ and is further normalized by $\boldsymbol{p}_c=\boldsymbol{p}'_c/\sum p'_i$. So the class with the largest probability is the prediction.

{\bfseries Least Euclidean Distance:}
Calculate the distances of test sample hidden representation $\boldsymbol{h}_t$ to the per-class mean hidden representations $\{\boldsymbol{\bar{h}}_1, \dots, \boldsymbol{\bar{h}}_c\}$, the class with the least distance is the prediction.

{\bfseries Classic LDA:}
A classic LDA classifier is trained by the hidden representation $\boldsymbol{H}$ of the training dataset $\boldsymbol{X}$, and then predicts the label of test sample hidden representation $\boldsymbol{h}_t$.

The purpose to compare the Least Euclidean Distance Predictor with the other two predictors is that the fact of nearly identical accuracy provides guarantee as well as confidence for us to improve DLDA from both the Euclidean and statistical viewpoints in Section \ref{section:RDLDA}.

\section{Regularized Deep Linear Discriminant Analysis} \label{section:RDLDA}

The inspiration leading to Regularized Deep Linear Discriminant Analysis(RDLDA) stemmed from the thought that if each dimension of the hidden representations not only separates a unique class but also utilizes the unoccupied space to distribute more classes, the discriminative ability of the model might rise. Though overlapping of some classes distributions might happen in one dimension, the unoverlapped distributions of these classes in other dimensions would make them clearly separated. The key idea is to strengthen the discriminative ability of each output dimension, and also keep them complement each other.

Each diagonal element of the within-class scatter matrix $\boldsymbol{S_W}$ is the variance of a single output dimension, which represents the internal information within that dimension; while each non-diagonal element is the covariance of a pair of dimensions, which reflects pairwise mutual relationship. In $\boldsymbol{S_W}$, the number of diagonal elements is much less than that of non-diagonal elements, which causes the dimension internal information overwhelmed by the pairwise mutual relationship. In other words, the small portion of diagonal elements and large portion of non-diagonal elements make DLDA DNN focus on training the cooperative discriminative ability of all the dimensions, while put less emphasis on training the separative capacity of single dimension. This explains the distribution pattern of DLDA hidden representations that discussed in Section \ref{section:Distribution Pattern}.

To strengthen the discriminative ability of each output dimension, $\boldsymbol{S_W}$ is regularized by the following formula to weaken the influence of non-diagonal elements that reflect dimensional pairwise mutual relationship.
\begin{equation}\label{eq:Regularized Sw}
\boldsymbol{S'_W} = \alpha \boldsymbol{S_W} + (1-\alpha)\mathbf{diag}(\boldsymbol{S_W})+\lambda \boldsymbol{I} \quad with \quad  0\le \alpha \le 1
\end{equation}
The operation $\mathbf{diag}(\boldsymbol{S_W})$ means leave the diagonal elements of $\boldsymbol{S_W}$ unchanged and set all the non-diagonal elements to zero. 

The partial derivative of the objective function with respect to hidden layer $\boldsymbol{H}$ becomes:
\begin{equation}\label{eq: partial derivative}
\frac{\partial}{\partial \boldsymbol{H}}\frac{1}{k}\sum_{i=1}^kv_i = 
\frac{1}{k}\sum_{i=1}^k\boldsymbol{e}_i^\top(\frac{\partial \boldsymbol{S_B}}{\partial \boldsymbol{H}} - v_i
\frac{\partial \boldsymbol{S'_W}}{\partial \boldsymbol{H}})\boldsymbol{e}_i
\end{equation}

Compared to the original $\boldsymbol{S_W}$, the diagonal elements of $\boldsymbol{S'_W}$ remain unchanged but all the non-diagonal elements are scaled down by a factor of $\alpha$. Note that when $\alpha=1$, $\boldsymbol{S'_W}=\boldsymbol{S_W}$, RDLDA is the same as DLDA. If $\alpha=0$, all the non-diagonal elements of $\boldsymbol{S'_W}$ are zero, in this case RDLDA DNN will concentrate on training the separative capacity of each single dimension while completely ignoring the coordination among dimensions. 

\begin{figure}[htp]
\begin{center}
\includegraphics[width=\textwidth]{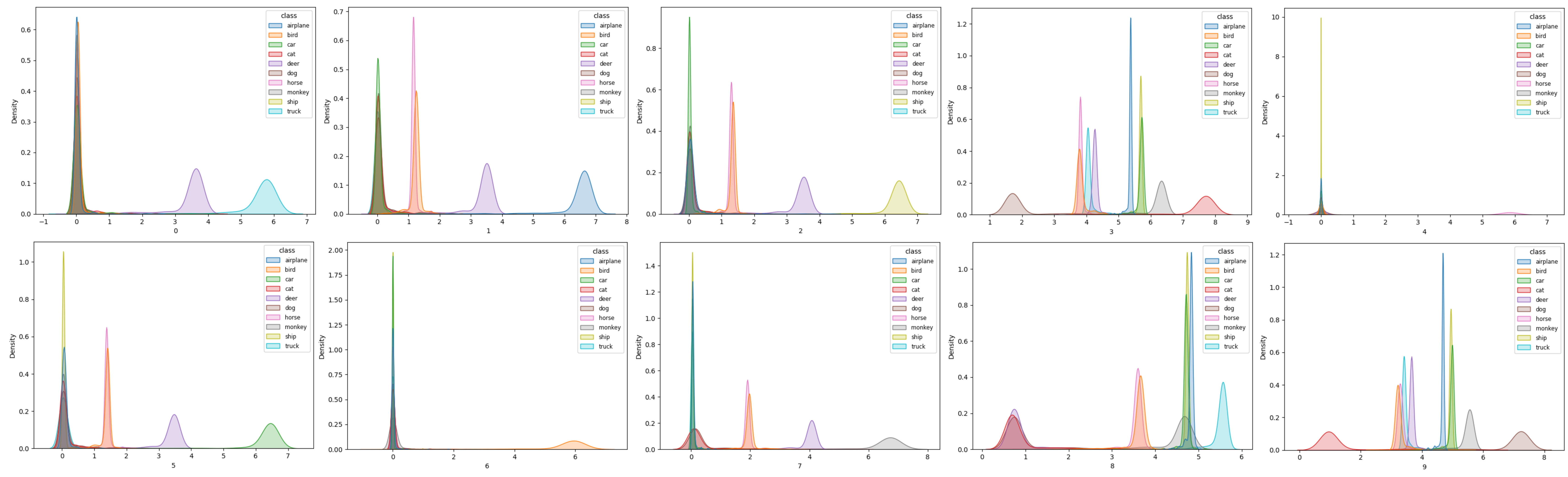}
\caption{Each subplot is the distribution of each output dimension of the RDLDA DNN with $\alpha=0$, which was trained 400 epochs by the STL-10 dataset.}\label{fig:STL10_train_output_0.0_kdeplot}
\end{center}
\end{figure}

To verify the above hypotheses, we regularized $\boldsymbol{S_W}$ by setting $\alpha=0$ and trained this new RDLDA DNN 400 epochs by the STL-10 dataset(the network structure and training parameters setting are detailed in the experiments section of this paper). The distribution of each output dimension is shown in each subplot of Fig.~\ref{fig:STL10_train_output_0.0_kdeplot} respectively, from which it can be seen that the separative ability of most dimensions is significantly improved.

Overemphasize cooperation at the cost of suppressing individual discretion, in this case $\alpha=1$, or overemphasize individual judgement at the expense of losing coordination,  in this case $\alpha=0$, might not be the best choice in regularizing $\boldsymbol{S_W}$. From the perspective of statistics, the process of $\alpha$ decreasing from $1$ to $0$ is also the process of the classes distributions moving from clustering to separating. The optimal value for $\alpha$ might be the moment when there is least overlapping of classes distributions. So in RDLDA, $\alpha$ is a hyperparmeter, of which the optimal value depends on the features of the dataset and the learning ability of the DNN on which RDLDA is based.

\section{Subclass RDLDA} \label{section:Subclass_RDLDA}
Two images of different classes but with similar background may be misclassified to the same class. To strengthen the discriminative ability in all the local space, instead of training the model discriminative ability in the global input space, we can split each class into serveral subclasses according to their similarity and train the model to seperate all the subclasses. K-means algorithm can be utilized to divide each class into k subclasses, but the distance of two samples cannot be simply measured by pixels difference in the input space, because an image generated by flipping horizontally may have large pixels difference with the original image. 

Autoencoder can extract hidden features of the input images and project them into the embedding representation space, within which, images with similar patterns or backgrounds are near each other. To learn similar features extraction ability and project samples to similar hidden space as RDLDA DNN, in our experiment design, the network architecture of the encoder layers is also similar with that of RDLDA DNN. The decoder layers are symmetric with the encoder layers. The last activation layer of encoder is Tanh function, so the range of each dimension in the embedding representation layer is $(-1, 1)$, which makes Euclidean distance an applicable measure of closeness. And the last activation layer of decoder is Sigmoid function, so the range of each output dimension is $(0, 1)$, which is the same as that of each input dimension. 

\begin{figure}[htp]
\begin{center}
\includegraphics[width=\textwidth]{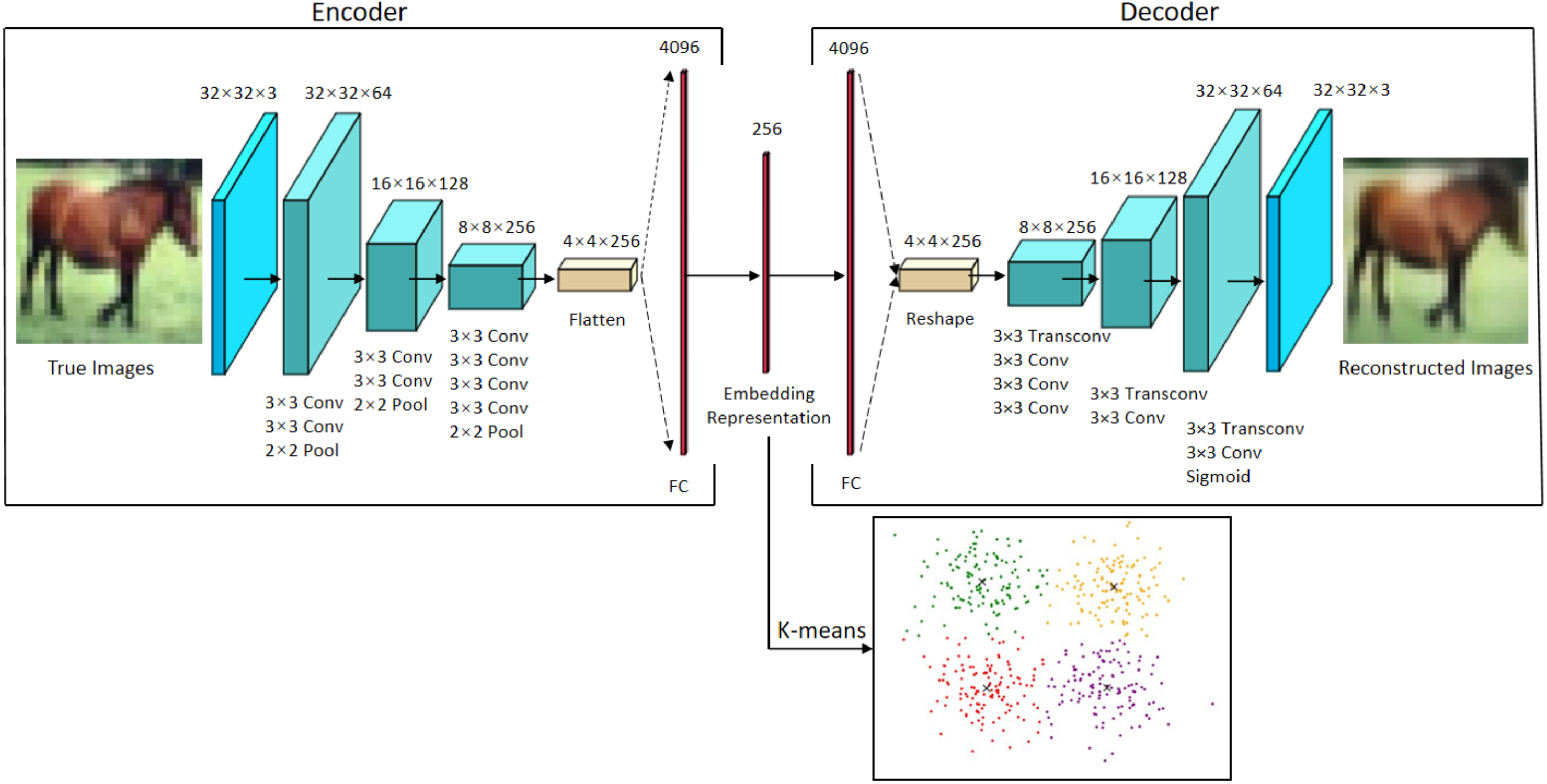}
\caption{Autoencoder network architecture and the reconstructed image, of which the original input image is from CIFAR-10 dataset.}\label{fig:Endecoder}
\end{center}
\end{figure}

Thus it is better to use K-means algorithm in the embedding representation space to divide each class into k subclasses. The subclasses splitting process and the autoencoder network architecture are illustrated in Fig.~\ref{fig:Endecoder}. And the training pipeline of Subclass RDLDA is summarized in \textbf{Algorithm 1}.

\begin{algorithm}
\caption{Subclass RDLDA}
\renewcommand{\algorithmicrequire}{\textbf{Input:}}
\renewcommand{\algorithmicensure}{\textbf{Output:}}

\begin{algorithmic}[1]
\REQUIRE training samples $\boldsymbol{x}_1, ..., \boldsymbol{x}_n$, labels $y_1, ..., y_n$, testing samples $\boldsymbol{x}_{t1}, ..., \boldsymbol{x}_{tm}$. 
\ENSURE predicted classes labels $\hat{y}_{t1}, ..., \hat{y}_{tm}$ of testing samples. \\
\textbf{Initialize:} number of endecoder iterations: $p$, number of subclasses per class: $k$, regularizing parameter: $\alpha$, number of epochs for RDLDA: $q$.

\FOR{$i=1$; $i<p$; $i++$ }  
\STATE {Forward the training samples $\boldsymbol{x}$ through the autoencoder network.}
\ENDFOR

\STATE Execute K-means clustering for each class $c \in C$ in the embedding representation space and assign subclass labels $s \in S$ to each training sample $\boldsymbol{x}$.

\STATE Compute $\boldsymbol{S'_W}$ according to Eq.~\eqref{eq:Regularized Sw}.

\FOR{$j=1$; $j<q$; $j++$ }  
\STATE {Optimize parameters of the RDLDA network by backpropagation according to the objective function: Eq.~\eqref{eq:eigenvalue loss}.}
\ENDFOR

\STATE Forward the testing samples $\boldsymbol{x}$ through the RDLDA network and predict subclasses labels $S \ni \hat{y'}_{t1}, ..., \hat{y'}_{tm}$.
\STATE Convert the subclasses labels $\hat{y'}_{t1}, ..., \hat{y'}_{tm}$ to classes labels $C \ni \hat{y}_{t1}, ..., \hat{y}_{tm}$.

\end{algorithmic}
\end{algorithm}

\section{Experiments and Results Analysis} \label{section:Experiments}  
\subsection{Experimental Method}
\begin{figure}[htp]
\begin{center}
\includegraphics[width=\textwidth]{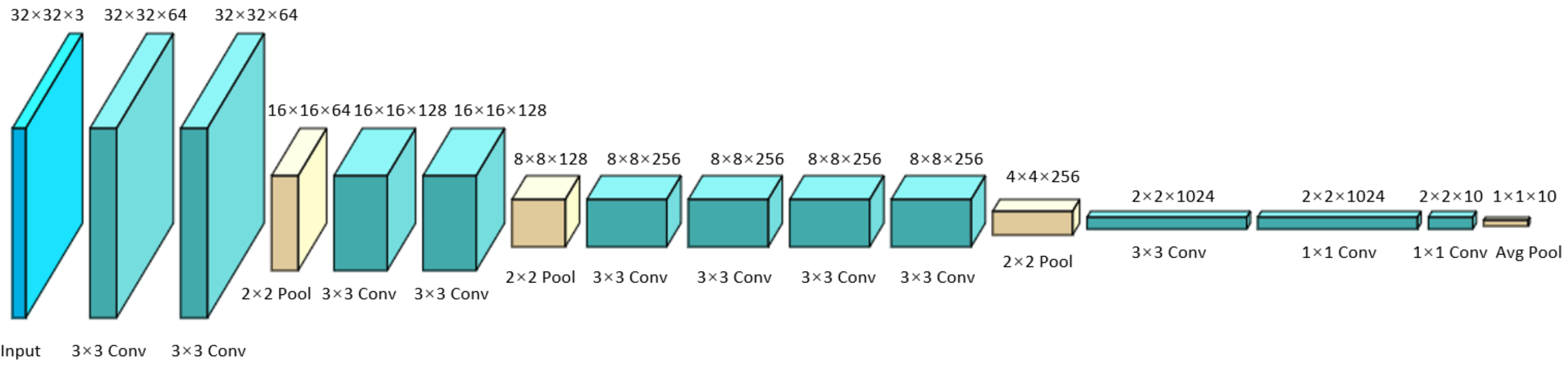}
\caption{DorferNet architecture.}\label{fig:DorfnetArch}
\end{center}
\end{figure}

To make a comparison with DLDA, the same network architecture and parameters in \cite{dorfer2015deep} were used in the following experiments. The network architecture renamed as \textit{DorferNet} is depicted in Fig.~\ref{fig:DorfnetArch} and outlined in Tab.~\ref{table:DorferNet}.

\begin{table}[htp]
\begin{center}
\begin{tabular}{c}

\hline
3 $\times$ 3 Conv(pad-1)-64-BN-ReLU \\ 
3 $\times$ 3 Conv(pad-1)-64-BN-ReLU \\ 
2 $\times$ 2 Max-Pooling + Drop-Out(0.25)  \\ 
\hline
3 $\times$ 3 Conv(pad-1)-128-BN-ReLU \\ 
3 $\times$ 3 Conv(pad-1)-128-BN-ReLU \\ 
2 $\times$ 2 Max-Pooling + Drop-Out(0.25)  \\ 
\hline
3 $\times$ 3 Conv(pad-1)-256-BN-ReLU \\ 
3 $\times$ 3 Conv(pad-1)-256-BN-ReLU \\ 
3 $\times$ 3 Conv(pad-1)-256-BN-ReLU \\ 
3 $\times$ 3 Conv(pad-1)-256-BN-ReLU \\ 
2 $\times$ 2 Max-Pooling + Drop-Out(0.25)  \\ 
\hline
3 $\times$ 3 Conv(pad-0)-1024-BN-ReLU \\ 
Drop-Out(0.5) \\ 
\hline
1 $\times$ 1 Conv(pad-0)-1024-BN-ReLU \\ 
Drop-Out(0.5) \\ 
\hline
1 $\times$ 1 Conv(pad-0)-10-BN-ReLU \\
Global-Average-Pooling \\ 
\hline

\end{tabular}
\caption{\cite{dorfer2015deep} named the above network architecture as \textit{OurNet}, in this paper it is renamed as \textit{DorferNet}. BN: Batch Normalization, ReLU: Rectified Linear Activation Function.}
\label{table:DorferNet}
\end{center}
\end{table}

In the experiments, all the networks were trained using SGD with Nesterov momentum. The initial learning rate was set to $0.1$ and was halved every 25 epochs, the momentum was fiexed at $0.9$, and the weight decay was 0.0001 on all trainable parameters. $\epsilon$ in Eq.~\eqref{eq:eigenvalue loss} was set to $1$, and $\lambda$ in Eq.~\eqref{eq:Regularized Sw} was set to 0.001. For RDLDA, the hyperparameter $\alpha$ in Eq.~\eqref{eq:Regularized Sw} was set to $0$, $0.1$, $0.2$, $0.3$, $0.4$, $0.5$, $0.6$, $0.7$, $0.8$, $0.9$ respectively. Note that RDLDA is DLDA when $\alpha=1$.

\cite{dorfer2015deep},  \cite{wang2015unsupervised}, \cite{wang2015deep} pointed out that mini-batch learning on distribution parameters was feasible if the batch-size was sufficiently large to be representative for the entire population. The available GPU(GTX1070 with 8GB memory) limited the batch size to 150 for the STL-10 dataset and 100 for the Pediatric Pneumonic Chest X-ray dataset. For the CIFAR-10 dataset, the batch size was set to 1000. The average number of images per class in a batch equals the batch size divided by the number of classes(subclasses). For the STL-10 dataset, if each class is divided into $k$ subclasses, with $k=2$ or $k=3$, the average number of images per subclass in a batch is only $7.5$ or $5$, which is too small to represent the whole subclass. And for the Pediatric Pneumonic Chest X-ray dataset, the single channel images are too simple and similar, applying K-means clustering on the dataset is meaningless. So \textbf{Algorithm 1} was only applied to the CIFAR-10 dataset in the following experiments. The number of training epochs was 400 for the STL-10 and CIFAR-10 dataset, and 200 for the Pediatric Pneumonic Chest X-ray dataset.

\subsection{Experimental Results on STL-10}
The STL-10( \cite{coates2011analysis} ) is an image recognition dataset with 10 classes, each class has 500 training images and 800 test images. It also contains 100000 unlabeled images for unsupervised learning, which are not used in the experiment. Images are 96 $\times$ 96 pixels, color. The original 5000 training images were split into a training set with 4000 images and a validation set with 1000 images for picking the best performing parametrization. Except normalization, no further processing was applied to the dataset.

\begin{table}[htp]
\begin{center}
\begin{tabular}{lccccc} 
\hline
Method & Hyperplanes & Euclidean & LDA & Softmax \\
 \hline
DorferNet RDLDA($\alpha=0$) & 81.1\% & 81.33\% & 80.75\% &    \\
DorferNet RDLDA($\alpha=0.1$) & 81.46\% & 81.46\% & 80.54\%  &   \\
DorferNet RDLDA($\alpha=0.2$) & \bfseries{81.51\%} & \bfseries{81.54\%} & \bfseries{81.7\%}  & \\
DorferNet RDLDA($\alpha=0.3$) & 81.06\% & 81.33\% & 81.31\%  &\\
DorferNet RDLDA($\alpha=0.4$) & 80.58\% & 81.09\% & 80.33\%  & \\
DorferNet RDLDA($\alpha=0.5$) & 80.77\% & 81.48\% & 80.97\%  & \\
DorferNet RDLDA($\alpha=0.6$) & 80.94\% & 81.45\% & 81.35\%  & \\
DorferNet RDLDA($\alpha=0.7$) & 80.17\% & 80.19\% & 80.26\%  & \\
DorferNet RDLDA($\alpha=0.8$) & 80.61\% & 80.69\% & 80.49\%  & \\
DorferNet RDLDA($\alpha=0.9$) & 80.74\% & 81.02\% & 81.51\%  & \\
 \hline
DorferNet DLDA & 80.16\% & 80.07\% & 80.06\%  &  \\
\hline
DorferNet CCE &  &  &  &  78.26\%\\
 \hline

\end{tabular}
\caption{Comparison of accuracy on the test set of STL-10. $\alpha$ is the hyperparameter in Eq.~\eqref{eq:Regularized Sw} for the regularization of the within-class matrix $\boldsymbol{S_W}$.}
\label{table:STL-10 experiment results}
\end{center}
\end{table}

It can be seen from Tab.~\ref{table:STL-10 experiment results} that the proposed regularization method shows fairly good performance, all the RDLDA networks performed better than the DLDA network, within which, RDLDA with $\alpha = 0.2$ achieved the highest accuracy, exceeding DLDA with around $1.5\%$.  And all the DLDA and RDLDA networks outperformed the CCE network, which proves LDA based deep learning objective is effective and competitive.

As a statistical model, LDA based DNN is more dependent on batch size than CCE based DNN. Technically, the larger the batch size, the more accurate the estimation of $\boldsymbol{S_W}$ and $\boldsymbol{S_B}$, the higher the test accuracy. 
Based on a more powerful GPU(Tesla K40) with larger memory(12GB), ( \cite{dorfer2015deep} ) set the batch size to 200, and in their paper DLDA achieved the accuracy of $81.16\%$. Although our experiments were run on a GPU(GTX1070) with less memory(8GB), which limited the batch size to 150, RDLDA($\alpha=0.2$) still surpassed DLDA with batch size of 200.

\subsection{Experimental Results on CIFAR-10}
The CIFAR-10( \cite{krizhevsky2009learning} ) is an image recognition dataset with 10 classes, each class has 5000 training images and 1000 test images. Images are 32 $\times$ 32 pixels, color. The original 50000 training images were split into a training set with 48000 images and a validation set with 2000 images for picking the best performing parametrization. Except normalization and random horizontal flipping, no further processing was applied to the dataset.

\begin{table}[htp]
\begin{center}
\begin{tabular}{lccccc} 
\hline
Method & Hyperplanes & Euclidean & LDA  & Softmax \\
 \hline
DorferNet RDLDA($\alpha=0$) & 91.86\% & 91.87\% & 91.84\%  &    \\
DorferNet RDLDA($\alpha=0.1$) & 91.78\% & 91.68\% & 91.64\%  &    \\
DorferNet RDLDA($\alpha=0.2$) & 92.06\% & 92.12\% & 91.85\%  & \\
DorferNet RDLDA($\alpha=0.3$) & 91.45\% & 91.43\% & 91.42\%  & \\
DorferNet RDLDA($\alpha=0.4$) & 91.15\% & 91.16\% & 91.13\%  &   \\
DorferNet RDLDA($\alpha=0.5$) & 91.51\% & 91.42\% & 91.39\%  &  \\
DorferNet RDLDA($\alpha=0.6$) & \bfseries{92.29\%} & \bfseries{92.28\%} & \bfseries{92.2\%}  &  \\
DorferNet RDLDA($\alpha=0.7$) & 91.77\% & 91.69\% & 91.72\%  &  \\
DorferNet RDLDA($\alpha=0.8$) & 91.51\% & 91.51\% & 91.5\%  &  \\
DorferNet RDLDA($\alpha=0.9$) & 91.28\% & 91.15\% & 91.16\%  &   \\
 \hline
DorferNet DLDA & 91.12\% & 91.04\% & 91.06\%  &   \\
\hline
DorferNet CCE &  &  &  &  90.94\%\\
 \hline

\end{tabular}
\caption{Comparison of accuracy on the test set of CIFAR-10. $\alpha$ is the hyperparameter in Eq.~\eqref{eq:Regularized Sw} for the regularization of the within-class matrix $\boldsymbol{S_W}$.}
\label{table:CIFAR-10 experiment results}
\end{center}
\end{table}

It can be seen from Tab.~\ref{table:CIFAR-10 experiment results} that the proposed regularization method also gained a satisfactory result, all the RDLDA networks performed better than the DLDA network and CCE network, within which, RDLDA with $\alpha = 0.6$ achieved the highest accuracy, exceeding DLDA by around $1.1\%$.

To test the effectiveness of \textbf{Algorithm 1}, we used the autoencoder network architecture as shown in Fig.~\ref{fig:Endecoder} and ran K-means in the embedding representation layer of 256 dimensions. It can been seen from Fig.~\ref{fig:Endecoder}, by utilizing this network architecture, the image was well reconstructed from the embedding representation layer, which proved the autoencoder network had well learned the ability to extract hidden features of the input images. The number of subclasse per class, $k$, is a hyperparameter. To attain the optimal value for subclasses dividing, $k$ was set to $2$, $3$, $4$ respectively. Hyperplanes were used for final classification in the following experiments.

\begin{table}[htp]
\begin{center}
\begin{tabular}{lcccc} 
\hline
Method & $k=2$ & $k=3$ & $k=4$  \\
 \hline
DorferNet RDLDA($\alpha=0$) & 92.24\% & 91.52\% & 90.16\%      \\
DorferNet RDLDA($\alpha=0.1$) & 92.46\% & 91.31\% & 90.33\%      \\
DorferNet RDLDA($\alpha=0.2$) & 92.37\% & 91.39\% & 90.21\%   \\
DorferNet RDLDA($\alpha=0.3$) & 92.31\% & 91.14\% & 89.93\%   \\
DorferNet RDLDA($\alpha=0.4$) & 92.22\% & 91.17\% & 89.97\%     \\
DorferNet RDLDA($\alpha=0.5$) & 92.35\% & 91.40\% & 89.62\%    \\
DorferNet RDLDA($\alpha=0.6$) & 92.43\% & 91.63\% & 90.03\%    \\
DorferNet RDLDA($\alpha=0.7$) & \bfseries{92.87\%} & 91.98\% & 90.14\%    \\
DorferNet RDLDA($\alpha=0.8$) & 92.28\% & 92.24\% & 89.2\%    \\
DorferNet RDLDA($\alpha=0.9$) & 92.14\% & 91.77\% & 90.19\%     \\
 \hline
DorferNet DLDA & 91.87\% & 91.02\% & 88.69\%     \\
\hline

\end{tabular}
\caption{Comparison of accuracy on the test set of CIFAR-10 for different $k$ values. $\alpha$ is the hyperparameter in Eq.~\eqref{eq:Regularized Sw} for the regularization of the within-class matrix $\boldsymbol{S_W}$.}
\label{table:CIFAR-10 subclasses experiment results}
\end{center}
\end{table}

It can be seen from Tab.~\ref{table:CIFAR-10 subclasses experiment results} that applying \textbf{Algorithm 1} and dividing each class into two subclasses outperformed the original algorithm by around $0.5\%$ for all $\alpha$, while dividing each class into three subclasses achieved almost the same result as the original algorithm for all $\alpha$. But further dividing each class into four subclasses performed worse than the original algorithm by around $1.5\%$ for all $\alpha$. The reason may explain why large $k$ values lead to lower accuracy is that under the same batch size, the larger the $k$ value is, the smaller the average subclass samples size in a batch, and the more biased the estimation of $\boldsymbol{S_W}$ and $\boldsymbol{S_B}$.

\section{Conclusion} \label{section:Conclusion} 
In this paper we propose RDLDA, a regularization method on within-class scatter matrix to strengthen the discriminative ability of each output dimension of DLDA. We evaluated the performance of the regularization method on three dataset, and the results suggest that the regularization method can successfully increase discriminability and raise accuracy. To further improve the discriminative ability of RDLDA in the local space, an algorithm named Subclass RDLDA is proposed, and its effectiveness was verified by our experiments.

The downside of LDA based DNN is that, as a statistical model, it is more dependent on large batch size than CCE based DNN. How to preprocess the original input data and design lighter DNN architectures to load more training samples in a batch or limited GPU memory will be our work in the next phase. In the future, we will also investigate the performance of LDA based DNN when it deals with unbalanced dataset and explore the possibilities to improve it.



\bibliography{acml21}






\end{document}